\documentclass{svproc}
\usepackage{url}
\usepackage[misc]{ifsym}

\usepackage{graphicx}        

\begin{document}
\mainmatter              
\title{Convolutional neural networks for crack detection on flexible road pavements}
\titlerunning{CNNs for crack detection on flexible road pavements}  
%
\author{Hermann Tapamo\inst{1} \and Anna Bosman\inst{2} \and
James Maina\inst{3} \and Emile Horak\inst{4}}
\authorrunning{Hermann Tapamo et al.} 
%
\institute{University of Pretoria, Department of Computer Science, Pretoria, South Africa,\\
\email{u14365597@tuks.co.za},
\and
University of Pretoria, Department of Computer Science, Pretoria, South Africa,\\
\email{anna.bosman@up.ac.za}
\and
University of Pretoria, Department of Civil Engineering, Pretoria, South Africa,\\
\email{james.maina@up.ac.za}
\and
University of Pretoria, Department of Civil Engineering, Pretoria, South Africa,\\
\email{emileh@global.co.za}
}
\maketitle              

\begin{abstract}
Flexible road pavements deteriorate primarily due to traffic and adverse environmental conditions. Cracking is the most common deterioration mechanism; the surveying thereof is typically conducted manually using internationally defined classification standards. In South Africa, the use of high-definition video images has been introduced, which allows for safer road surveying. However, surveying is still a tedious manual process. Automation of the detection of defects such as cracks would allow for faster analysis of road networks and potentially reduce human bias and error. This study performs a comparison of six state-of-the-art convolutional neural network models for the purpose of crack detection. The models are pretrained on the ImageNet dataset, and fine-tuned using a new real-world binary crack dataset consisting of 14000 samples. The effects of dataset augmentation are also investigated. Of the six models trained, five achieved accuracy above 97\%. The highest recorded accuracy was 98\%, achieved by the ResNet and VGG16 models. The dataset is available at the following URL: \url{https://zenodo.org/record/7795975}
\end{abstract}
\section{Introduction}
\label{sec:1}

Globally, the most popular form of transport is the motor vehicle. Consequently, public roads experience significant traffic load, which, coupled with the environmental and climatic conditions, are the primary contributors to rapid road surface deterioration~\cite{Shahbazi2021}. 

A pavement management system (PMS) provides a systematic technique for recording and analysing pavement surface conditions of the road network, and enables personnel in charge to make data-driven maintenance decisions. An efficient monitoring strategy facilitates the development of a maintenance schedule that would help to significantly reduce life-cycle maintenance costs~\cite{Gopalakrishnan2017}. The performance of the PMS depends on the quality of the pavement surface condition data recorded. Inspection methods may be divided into three main categories; manual, semi-automated, and automated. The manual inspection is carried out in the field, it is laborious and unsafe. In semi-automated inspections, on-vehicle sensors/cameras record the pavement surface conditions to be analysed manually later. Automated inspection consists of on-vehicle sensors coupled with distress detection algorithms, able to determine pavement conditions automatically. Automated inspection has not yet been widely adopted~\cite{Chatterjee2018}, despite its great potential.

The road surface condition inspection and data collection is still a predominantly manual process conducted by field personnel~\cite{Jenkins2018}. Apart from its time-consuming nature, personnel risk exposure, low detection efficiency and traffic congestion during inspection~\cite{Wang2020}. Manual approach introduces human subjectivity and error, resulting in variability in how the road data is interpreted from person to person~\cite{Kaseko1993}. Over the years, there have been numerous attempts to embed various technologies into the road inspection process, with the primary aims of counteracting human variability, improving the survey execution speed, and mitigating personnel on-site risk exposure.

Research efforts aimed at introducing a level of automation to this field are focused on the application of image processing and computer vision technologies~\cite{Cafiso2017,Jenkins2018}. Advancements in deep learning and its increased accessibility have led to adoption of the deep learning methods in multiple fields, including pavement inspection. However, no comprehensive comparison of the effectiveness of the various deep learning methods exists to date. The availability of realistic high-definition road crack datasets also remains limited.

One of the most common defects which pavements are susceptible to is cracks, which develop over time primarily due to one or more of the following factors: repeated traffic loading, hostile environmental or climatic conditions and construction quality~\cite{Lau2020}. These factors contribute to the pavement ageing process and impact its structural integrity \cite{Konig2019}. This study considers cracks as a reliable visual indicator of pavement surface condition, and therefore explores the viability of using convolutional neural networks (CNNs) together with transfer learning to detect pavement cracks. The novel contributions of this study are summarised as follows:
\begin{itemize}
   \item A new binary crack image dataset is collected, consisting of 14000 samples used to train and test the various models evaluated.
   \item Six different state-of-the-art CNN models are applied to the collected dataset, initially by only training a binary classifier on top of pretrained model layers, and subsequently fine-tuning several top layers. A comparative study is performed to determine the effectiveness of transfer learning and fine-tuning.
   \item The comparative study is further enhanced by investigating the influence of data augmentation on the training process. 
   \item Based on all of the above, a recommendation is made as to which of the current state-of-the-art models seems to be the most appropriate for the chosen application domain.
\end{itemize}
The rest of the paper is structured as follows: Section~\ref{sec:bg} briefly outlines the background of this study. Section~\ref{sec:method} details the methodology. Section~\ref{sec:results} presents and discusses the results. Section~\ref{sec:conclusion} concludes the paper.

\section{Background}\label{sec:bg}
South Africa possesses the longest interconnected road network in sub-Saharan Africa, with a total length of 750,811 kilometres \cite{SARoad2021}. The South African National Roads Agency (SANRAL) is responsible for the management, maintenance and expansion of the major highways. Regional departments and local municipalities oversee the provincial/regional routes and smaller urban roads, respectively. The state of roads in South Africa is generally regarded as poor, 54\% of unpaved roads are in poor to very poor condition, and the same is said of 30\% of paved roads in South Africa \cite{SARoad2021}. Embedding deep learning into road surveying, which is currently performed manually, can make the process more efficient, less laborious and capable of producing more reliable and consistent records.

This study applies CNNs for the purpose of crack detection on a real-world dataset collected in South Africa. A CNN is a class of artificial neural network notably useful in computer vision tasks such as object recognition~\cite{OShea2015}, due to its ability to extract high-level features from images and thereby reliably recognize various objects after the model is trained. CNNs eliminate the need for manual feature extraction, which primitive image processing methods rely on, and instead learns the image features and characteristics directly from the input data; pre-processing efforts are therefore significantly reduced.

Transfer learning is a machine learning method that makes use of a previously trained model as a starting point for training a model on a new task~\cite{Zhuang2021}. It leverages past knowledge to extract valuable features from the new dataset being used. If successful, it should allow for faster model training and provide a performance boost. Training from scratch is compute-intensive and generally requires large datasets to achieve acceptable performance.

\section{Methodology}\label{sec:method}

\subsection{Dataset}\label{subsec:data} 
\begin{figure}
\includegraphics[width=\textwidth]{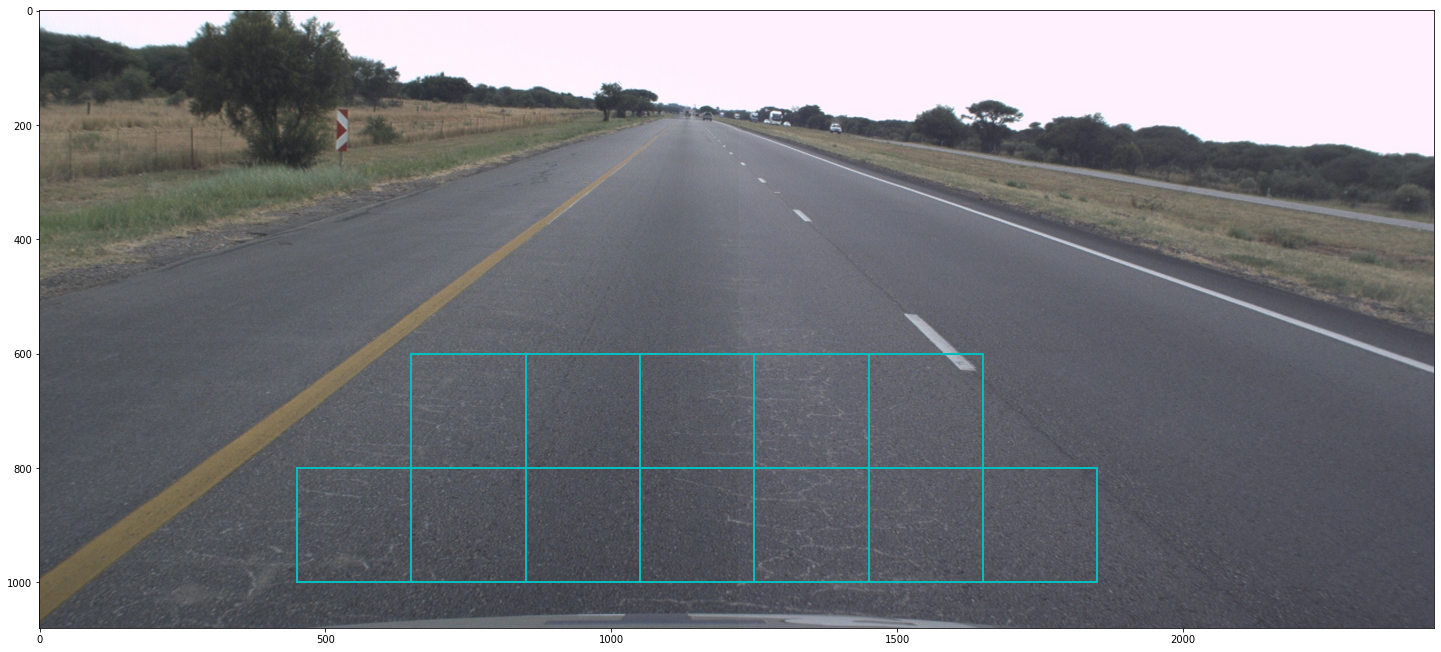}
\caption{Example image captured by MFV superimposed with sample regions extracted for the dataset.} \label{data_prep}
\end{figure}

The images were captured using a multi-functional vehicle (MFV) with an adapted high-definition wide angle camera \cite{horakmaina}. Each original image has a resolution of 2440x1080 pixels. At each capture point, there is a limited distance to which relevant image data can be collected, due to the camera focus and resolution. The images are collected on the slow lane, which generally carries more load due to heavier vehicles spending more time on it, consequently making the slow lane more susceptible to deterioration. Twelve square images of 200x200 pixels were extracted from each original image as shown in Figure \ref{data_prep}.

\begin{figure}[t]\centering
\includegraphics[width=\textwidth]{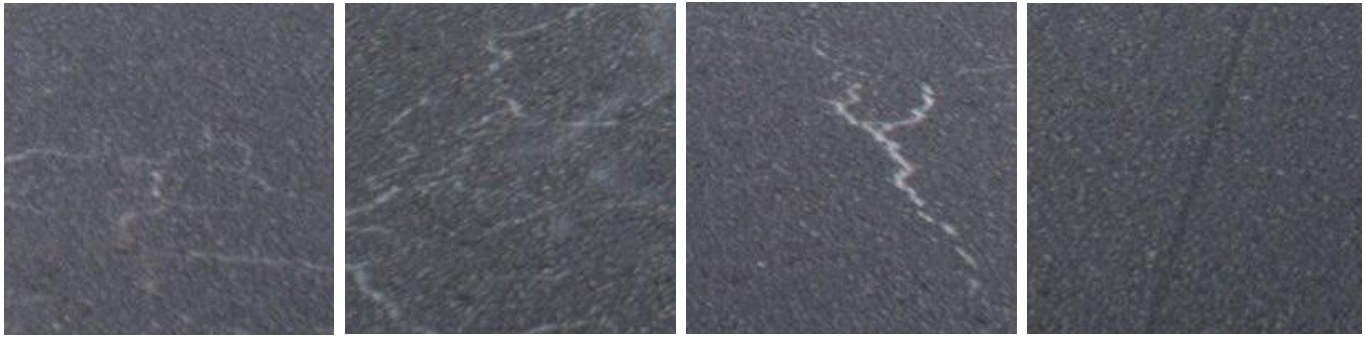}
\caption{Example of positive samples (cracks present).} \label{positive_samples}
\end{figure}

\begin{figure}[t]\centering
\includegraphics[width=\textwidth]{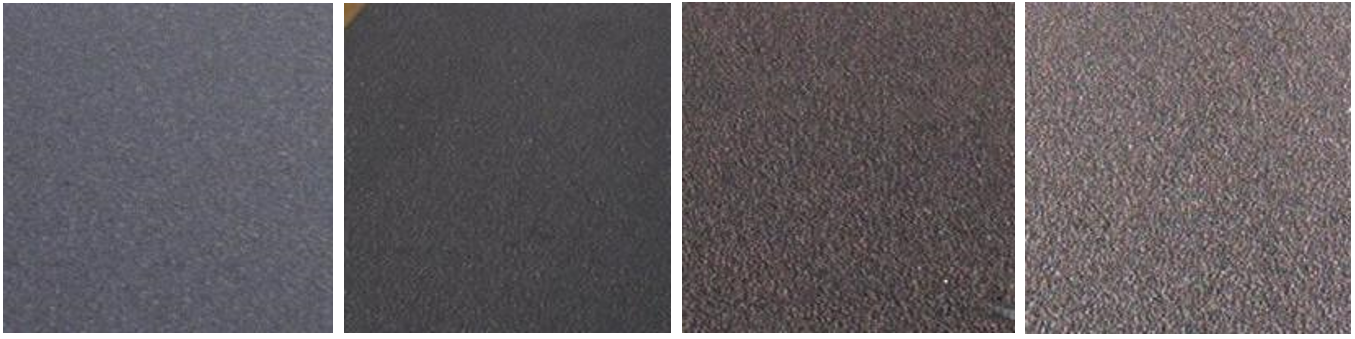}
\caption{Example of negative samples (cracks absent).} \label{negative_samples}
\end{figure}

A dataset consisting of 14000 images was built to train the models, split evenly between positive (cracks present) and negative samples (cracks absent), examples of each shown in Figures \ref{positive_samples} and \ref{negative_samples}, respectively. The 14000 images are a combination of the training data (11200 images), validation data (1400 images) and test data (1400 images). The training data is used to train the machine learning model. The validation data is held back during the training and used to provide an unbiased model performance update throughout the training process, useful for tuning the model hyperparameters. The test data, as with the validation data, is also unseen during model training, used once training is complete to compare the final models.
Each of these three data components is split evenly between positive and negative images, where positive images are samples consisting of cracks, and negative images are samples free of cracks. These datasets have been manually prepared to be used for model training. Experiments are conducted on the original dataset, as well as an augmented dataset, where the original data is expanded by flipping and rotating the images.

The dataset is available online: \url{https://zenodo.org/record/7795975}.

\subsection{Deep Learning Models}\label{subsec:models} Transfer learning was applied to six image classification models pre-trained on the ImageNet dataset. The two classes, in this case, are; positive (crack present) and negative (crack absent), making this a binary classification task.

The following CNN models were evaluated:
\begin{enumerate}
\item EfficientNet \cite{Tan2019} – a model that proposes a new scaling method which scales all dimensions using an effective compound coefficient. EfficientNet has been found to achieve a state-of-the-art performance compared to other CNNs with a significantly smaller architecture.
\item Inception V3 - a state-of-the-art CNN model based on the original paper by Szegedy et al.~\cite{Szegedy2016}, which performs convolutions on various scales (called inception modules) throughout the architecture.
\item Xception \cite{Chollet2017} - a variation on Inception V3, where inception modules are substituted for depth-wise separable convolutions. The architecture has an identical number of parameters as Inception V3.
\item MobileNet \cite{Howard2017} – efficient CNN model designed for mobile and embedded vision applications, based on a streamlined architecture which makes use of convolutions separable by depth.
\item ResNet \cite{He2016} – ResNet, or residual network, is a CNN architecture that adds residual connections between hidden states, allowing for better error propagation duing training.
\item VGG16 \cite{Simonyan2015} - a state-of-the-art CNN architecture consisting of 16 convolutional layers.
\end{enumerate}


The methodology adopted for the training of the models consists of two phases. In the first phase, pre-trained models are adapted to the dataset by only training the final fully-connected layer. In the second phase, further fine-tuning is performed through the unfreezing of several top layers and reducing the rate of learning to prevent overfitting. 
The proportion of top layers unfrozen for fine-tuning out of the total layers in each architecture was approximately 25\%.
A total of 80 epochs were completed for each model. Fine-tuning (second phase) was started at epoch 60.
A batch size of 32 was used in the experiments.

The cracks that can be detected include: block, longitudinal, transverse, crocodile and combinations of the aforementioned types, however, these are all detected as positive, reducing the problem to binary classification. The model was trained using samples up to a degree 1 type crack as per TMH9~\cite{TMH9FP}. Each model was trained with and without augmentation. This results in two models trained per architecture, thus twelve models in total. In order to draw inference on the consistency and reliability of the performance achieved, each model has been trained 5 times and all reported values are averages across the 5 runs, unless stated otherwise.

\section{Results}\label{sec:results}

\begin{figure}[b]
\includegraphics[width=\textwidth]{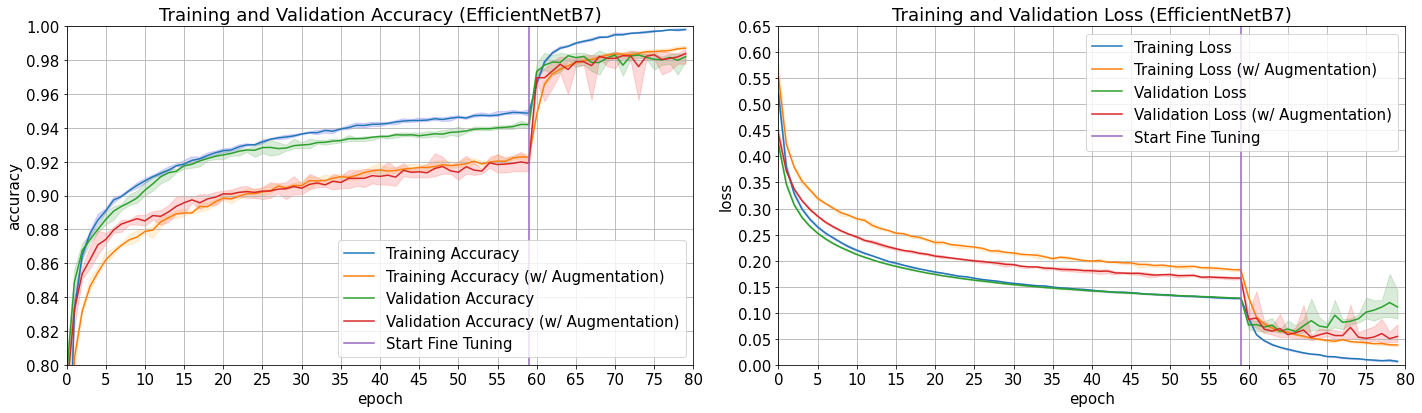}
\caption{EfficientNetB7 Accuracy/Loss vs Epoch.} \label{efficientnetb7}
\end{figure}

The EfficientNetB7 model converged to lower accuracy values and higher loss values for the version trained with data augmentation compared to that without, prior to fine-tuning, as seen in Figure \ref{efficientnetb7}. The range of values, represented by the shaded region, shows generally narrow ranges for all metrics prior to fine-tuning, indicative of training stability. Once fine-tuning is initialized, there is a notable increase in accuracy across the remaining epochs. The model trained with data augmentation exhibits superior stability over that without data augmentation, as shown in the similar training and validation accuracy and loss obtained at each epoch. A divergence in the training and validation loss and accuracy occurs in the model trained without augmentation, indicating overfitting.

\begin{figure}[t]
\includegraphics[width=\textwidth]{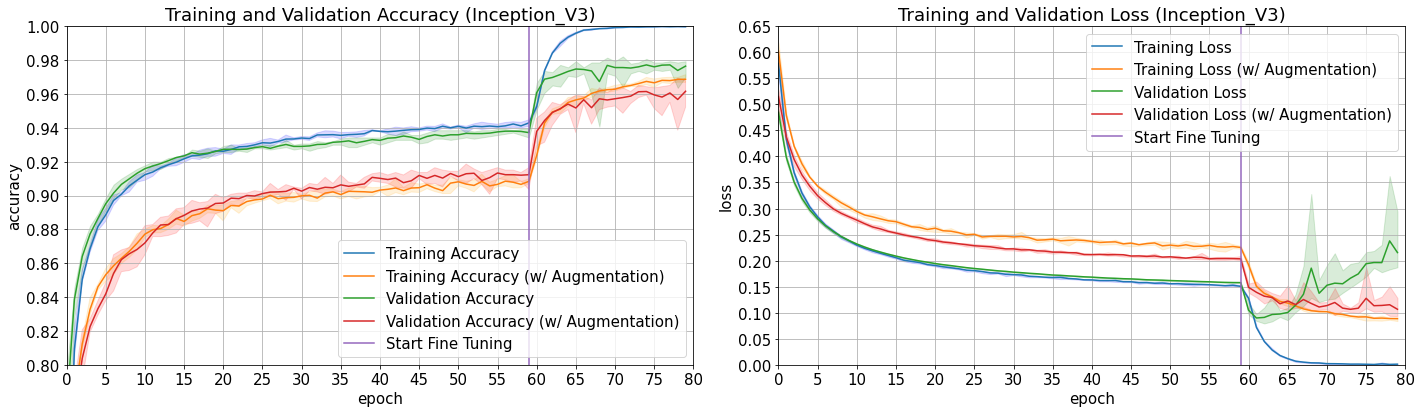}
\caption{Inception V3 Accuracy/Loss vs Epoch.} \label{inception_v3}
\end{figure}

\begin{figure}[t]
\includegraphics[width=\textwidth]{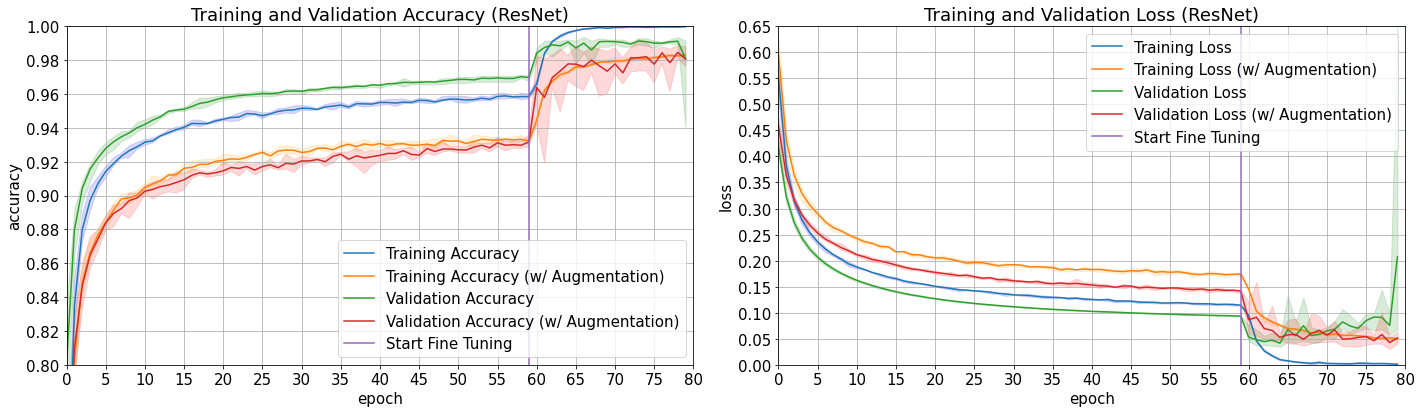}
\caption{ResNet Accuracy/Loss vs Epoch.} \label{resnet}
\end{figure}

\begin{figure}[t]
\includegraphics[width=\textwidth]{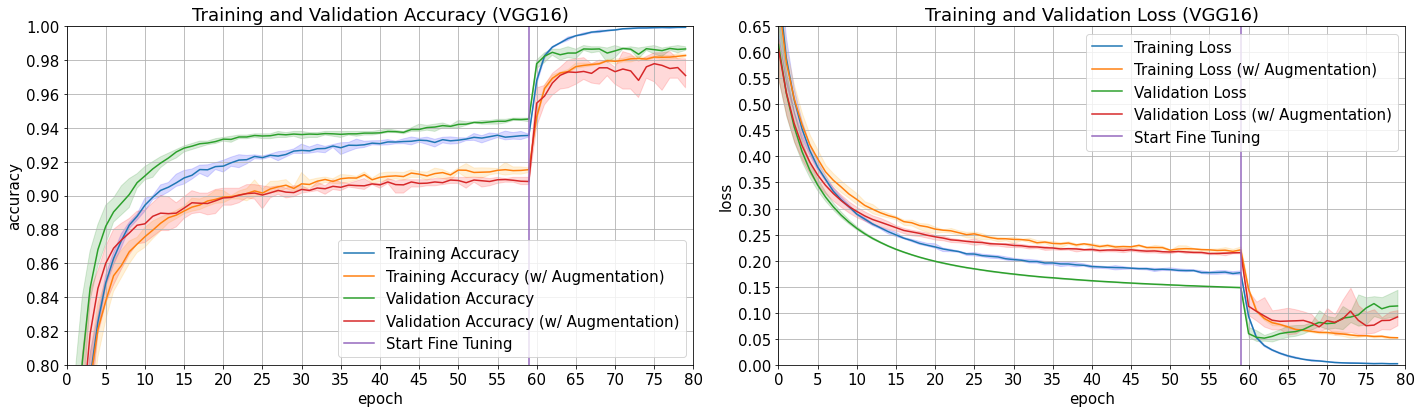}
\caption{VGG16 Accuracy/Loss vs Epoch.} \label{vgg16}
\end{figure}

The aforementioned trends are repeated in Figure \ref{inception_v3} for the Inception V3 model, Figure \ref{resnet} for the ResNet model, and Figure \ref{vgg16} for the VGG16 model.

The most significant divergence in the training and validation accuracy and loss occurs with the Inception V3 model.

\begin{figure}
\includegraphics[width=\textwidth]{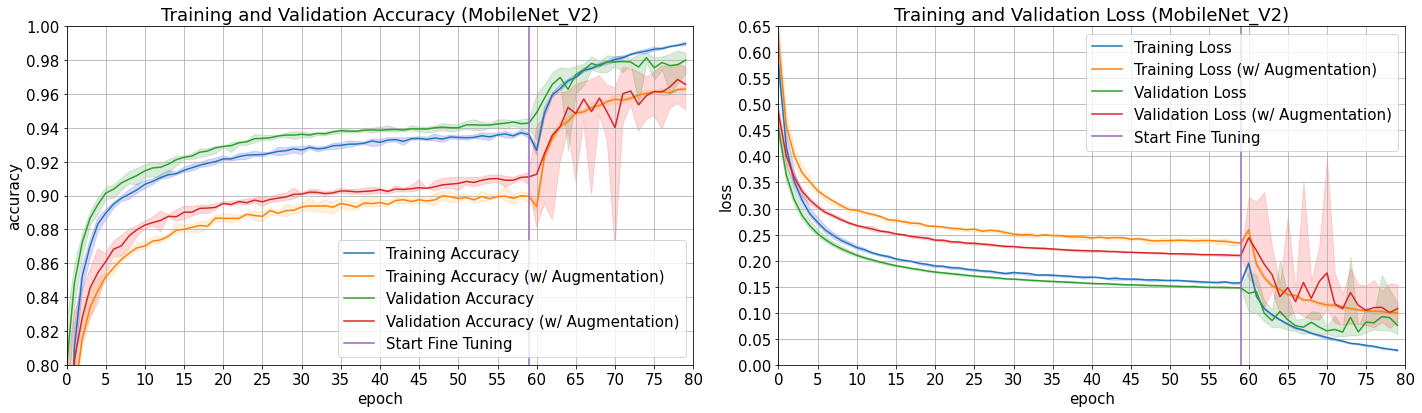}
\caption{MobileNet V2 Accuracy/Loss vs Epoch.} \label{mobilenet_v2}
\end{figure}

The MobileNet V2 model training values shown in Figure \ref{mobilenet_v2} show improvement in performance with the initialization of fine-tuning. The training and validation values of both versions (with and without augmentation) remain similar across all epochs, however these are based on the average values across the 5 separate training iterations. The shaded regions show that the MobileNet V2 validation accuracy and loss change significantly across the iterations between epoch 60 and 80, where fine-tuning is enabled. This suggests inferior robustness during training. It may therefore be inferred that the MobileNet V2 model has the poorest repeatability amongst all models evaluated when comparing the loss and accuracy ranges with that of the other models.

The trends exhibited in training the Xception model are distinctly different in the epochs prior to fine-tuning being initialized, with the data augmentation version reporting significantly different values of validation and training accuracy and loss, however upon the initialization of fine-tuning, the values for these metrics began to converge.

\begin{figure}[t]
\includegraphics[width=\textwidth]{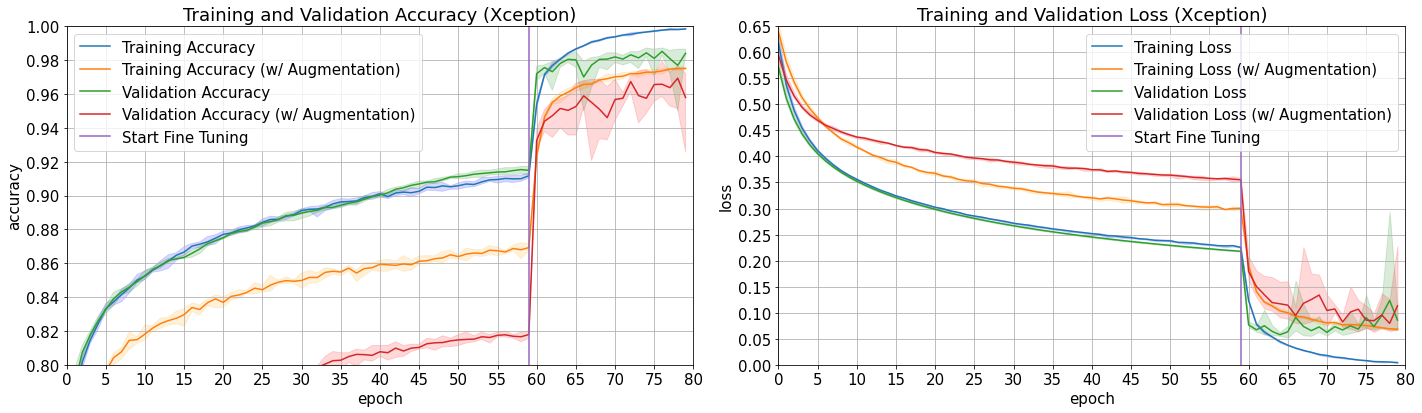}
\caption{Xception Accuracy/Loss vs Epoch.} \label{xception}
\end{figure}

The significant increase in performance achieved across the board upon the initialization of fine tuning confirms the theory hypothesized; the feature representation in the base model has been made more relevant and better suited for the task of crack classification. The introduction of fine-tuning, however, does also bring about generally higher variability in the validation accuracy and loss across the epochs. This suggests notable compromise to the repeatability in performance across multiple runs.

\begin{figure}[t]\centering
\includegraphics[width=0.8\textwidth]{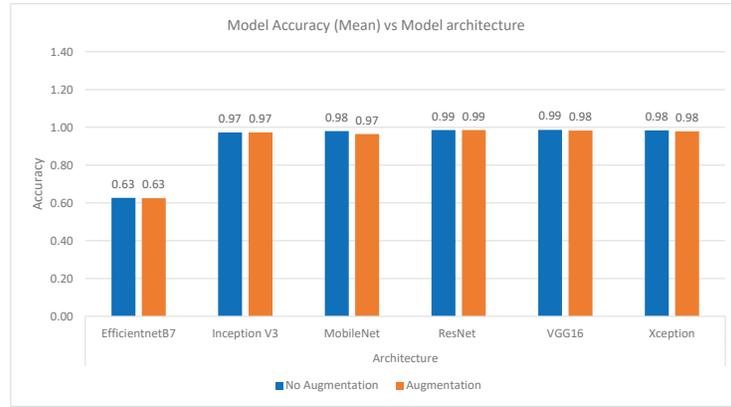}
\caption{Model accuracy on the test set with and without augmentation.} \label{model_accuracy}
\end{figure}

The test accuracy results are shown in Figure \ref{model_accuracy}.
The models performed well on the test set, with the exception of EfficientNetB7 model. This may be due to several reasons; the scaling method proposed in the architecture may not have been well suited for the model to learn the relevant crack features, the training and validation datasets may have had samples with similar features, and therefore overfitting occurred, resulting in poor performance when exposed to the unseen test dataset with distinctly different features. As a result of this, the EfficientnetB7 model achieved an accuracy of 0.63, while the other models achieved accuracy between 0.97 and 0.99.

Table \ref{performance_metrics} shows the model performance metrics which includes the mean F1 score calculated using precision and recall. The EfficientNetB7 model registers the highest precision values, with a score of 1, due to the fact that the model does not register any false positives. However, a disproportionate number of samples are classified as negative, as seen in Table \ref{confusion_metrics} (approximately 87\%). This is what results in the model having a relatively poor accuracy (0.626/0.625) and F1-score (0.402/0.400).

\begin{table}[t]
\caption[A short table caption]{Model performance metrics}
\resizebox{\textwidth}{!}{%
\begin{tabular}{c c c c c c c c c c c c c}
        \cline{2-13}
                            & \multicolumn{2}{c}{EfficientNetB7} & \multicolumn{2}{c}{Inception} & \multicolumn{2}{c}{MobileNet} & \multicolumn{2}{c}{ResNet} & \multicolumn{2}{c}{VGGNet} & \multicolumn{2}{c}{Xception} \\\cline{2-13}
                            & Non-Aug & Aug & Non-Aug & Aug & Non-Aug & Aug & Non-Aug & Aug & Non-Aug & Aug & Non-Aug & Aug \\\cline{1-13}
        Precision                 & 1.0000* & 1.0000* & 0.9874 & 0.9914 & 0.9921 & 0.9787 & 0.9933 & 0.9933 & 0.9894 & 0.9936 & 0.9871 & 0.9904 \\
        Recall (Mean)             & 0.2529 & 0.2503 & 0.9583 & 0.9534 & 0.9686 & 0.9509 & 0.9774 & 0.9786 & 0.9831 & 0.9711 & 0.9800 & 0.9666 \\
        Accuracy (Mean)           & 0.6264 & 0.6251 & 0.9730 & 0.9726 & 0.9804 & 0.9650 & 0.9854 & 0.9860 & 0.9863 & 0.9824 & 0.9836 & 0.9786 \\
        Accuracy (SD)             &  0.0210 & 0.0148 & 0.0020 & 0.0021 & 0.0021 & 0.0044 & 0.0011 & 0.0034 & 0.0015 & 0.0016 & 0.0015 & 0.0041 \\
        F1-Score (Mean)           & 0.4022 & 0.3997 & 0.9726 & 0.9720 & 0.9802 & 0.9645 & 0.9853 & 0.9859 & 0.9862 & 0.9822 & 0.9835 & 0.9783 \\
        F1-Score (SD)             & 0.0536 & 0.0376 & 0.0021 & 0.0023 & 0.0021 & 0.0046 & 0.0011 & 0.0034 & 0.0015 & 0.0017 & 0.0015 & 0.0044 \\
        Finetune Accuracy Boost   & 0.0410 & 0.0640 & 0.0390 & 0.0480 & 0.0380 & 0.0580 & 0.0210 & 0.0530 & 0.0420 & 0.0680 & 0.0700 & 0.1510 \\
        Finetune Loss Reduction   & 0.0600 & 0.1160 & 0.0680 & 0.0970 & 0.0850 & 0.1100 & 0.0520 & 0.0990 & 0.0970 & 0.1410 & 0.1600 & 0.2760 \\
        \cline{1-13}
    \end{tabular}} \label{performance_metrics}
\end{table}

The highest performance increase experienced once fine-tuning was enabled was from the data augmentation version of the Xception model both in accuracy (increase) and loss (decrease), with respective values of 0.151 and 0.276. The best accuracy and F1 score, both being 0.986, are achieved by the version of ResNet trained with data augmentation, and that of VGGNet without data augmentation. The standard deviations for the accuracy values are 0.003 and 0.002, and for the F1-scores 0.003 and 0.002, respectively. The standard deviations of both the F1-scores and accuracy of all models, as shown in Table \ref{performance_metrics}, are extremely low, ranging between 0.001 and 0.005, apart from EfficientNetB7, ranging between 0.015 and 0.054. It can therefore be inferred that, apart from the EfficientNetB7 models, the performance of the models reported may be deemed reliable and repeatable.

Table \ref{confusion_metrics} shows the mean values of the confusion matrix inputs. A notable takeaway is the difference in the number of false positives and false negatives. All models have more than double (in some instances more than triple) the number of false negatives as the number of false positives. This infers that the probability for a model to detect a false negative is more than double the probability of it detecting a false positive. This is not a favourable statistic; therefore, an argument can be made that of the two best performing models, that which is less likely to detect a false negative may be deemed superior. The VGGNet model version without augmentation is therefore the best trained model.

\begin{table}
\caption[A short table caption]{Confusion matrix metrics}
\resizebox{\textwidth}{!}{%
\begin{tabular}{c c c c c c c c c c c c c}
        \cline{2-13}
                            & \multicolumn{2}{c}{EfficientNetB7} & \multicolumn{2}{c}{Inception} & \multicolumn{2}{c}{MobileNet} & \multicolumn{2}{c}{ResNet} & \multicolumn{2}{c}{VGGNet} & \multicolumn{2}{c}{Xception} \\\cline{2-13}
                            & Non-Aug & Aug & Non-Aug & Aug & Non-Aug & Aug & Non-Aug & Aug & Non-Aug & Aug & Non-Aug & Aug \\\cline{1-13}
        True Positive   & 177 & 175 & 671 & 667 & 678 & 666 & 684 & 685 & 688 & 680 & 686 & 677 \\
        False Positive  & 0 & 0 & 9 & 6 & 5 & 15 & 5 & 5 & 7 & 4 & 9 & 7\\
        True Negative   & 700 & 700 & 691 & 694 & 695 & 685 & 695 & 695 & 693 & 696 & 691 & 693 \\
        False Negative  & 523 & 525 & 29 & 33 & 22 & 34 & 16 & 15 & 12 & 20 & 14 & 23 \\
        \cline{1-13}
    \end{tabular}} \label{confusion_metrics}
\end{table}

\section{Conclusion}\label{sec:conclusion}
This study presented a comparison between a selection of state-of-the-art CNN models applied to a real-life crack detection dataset collected in South Africa. The work presented herein shows great potential for the use of CNN architectures in conjunction with transfer learning for binary crack classification. Of all the models trained, VGGNet without data augmentation yielded the best performance overall. The dataset compiled consisted of only 14000 images, hence an attempt was made to compensate for it with data augmentation. The boost in performance yielded however, was minimal, but not detrimental to the performance. In future, additional data augmentation techniques may be considered. The aid of machine learning models would yield multiple benefits to road inspections. The research herein shows potential, if not for a fully autonomous road survey program, then at a minimum to assist surveyors in the consistent detection of road defects.

\bibliographystyle{splncs03}
\bibliography{crack_ref}

\end{document}